\documentclass{article} 
\usepackage{iclr2024_conference,times}

\usepackage{amsmath,amsfonts,bm}

\def\eqref#1{equation~\ref{#1}}

\def\1{\bm{1}}

\DeclareMathAlphabet{\mathsfit}{\encodingdefault}{\sfdefault}{m}{sl}
\SetMathAlphabet{\mathsfit}{bold}{\encodingdefault}{\sfdefault}{bx}{n}

\usepackage{url}

\usepackage[pagebackref,breaklinks,colorlinks]{hyperref}

\usepackage[utf8]{inputenc} 
\usepackage[T1]{fontenc}    

\usepackage{booktabs}
\usepackage{graphicx}
\usepackage{tablefootnote}
\usepackage{amsmath}
\usepackage{amssymb}
\usepackage{bbm}

\usepackage{cite}
\usepackage{amsmath,amssymb,amsfonts}
\usepackage{algorithmic}
\usepackage{textcomp}
\usepackage{subcaption}
\usepackage{tabularx}
\usepackage{booktabs}       
\usepackage[normalem]{ulem}
\useunder{\uline}{\ul}{}
\usepackage{amsmath}
\usepackage{amssymb}
\usepackage{bm}
\usepackage{color}
\usepackage{xcolor}
\usepackage{algorithm}
\usepackage{multirow}
\usepackage{xspace}

\usepackage{enumitem}

\usepackage{newtxtext}
\usepackage[capitalize]{cleveref}
\crefname{section}{Sec.}{Secs.}
\Crefname{section}{Section}{Sections}
\Crefname{table}{Table}{Tables}
\crefname{table}{Tab.}{Tabs.}

\definecolor{demphcolor}{RGB}{144, 144, 144}
\definecolor{mygray}{gray}{0.4}
\definecolor{lightgray}{rgb}{0.9, 0.9, 0.9}
\definecolor{deepgreen}{RGB}{0,100,0}

\hypersetup{%
  citecolor=teal
}
\hypersetup{linkcolor = black}
\newcommand{\modelname}{mPLUG-Owl3\xspace}

\title{mPLUG-Owl3: Towards Long Image-Sequence Understanding in Multi-Modal Large Language Models}

\iclrfinalarxiv 
\begin{document}

\author{
}

\maketitle

\begin{center}
   \centering
   \vspace{-16mm}
   \textbf{Jiabo Ye\footnote{Equal contribution}\qquad Haiyang Xu\footnotemark[1]\qquad Haowei Liu\qquad Anwen Hu \qquad Ming Yan\footnote{Corresponding author}} \\ 
    \textbf{Qi Qian\qquad Ji Zhang\qquad Fei Huang \qquad Jingren Zhou} \\
    Alibaba Group \\
    {\tt\small \{yejiabo.yjb, shuofeng.xhy, ym119608\}@alibaba-inc.com} 
   
   \url{https://github.com/X-PLUG/mPLUG-Owl}
\end{center}

\begin{figure*}[!ht]
    \centering
    \includegraphics[width=\textwidth]{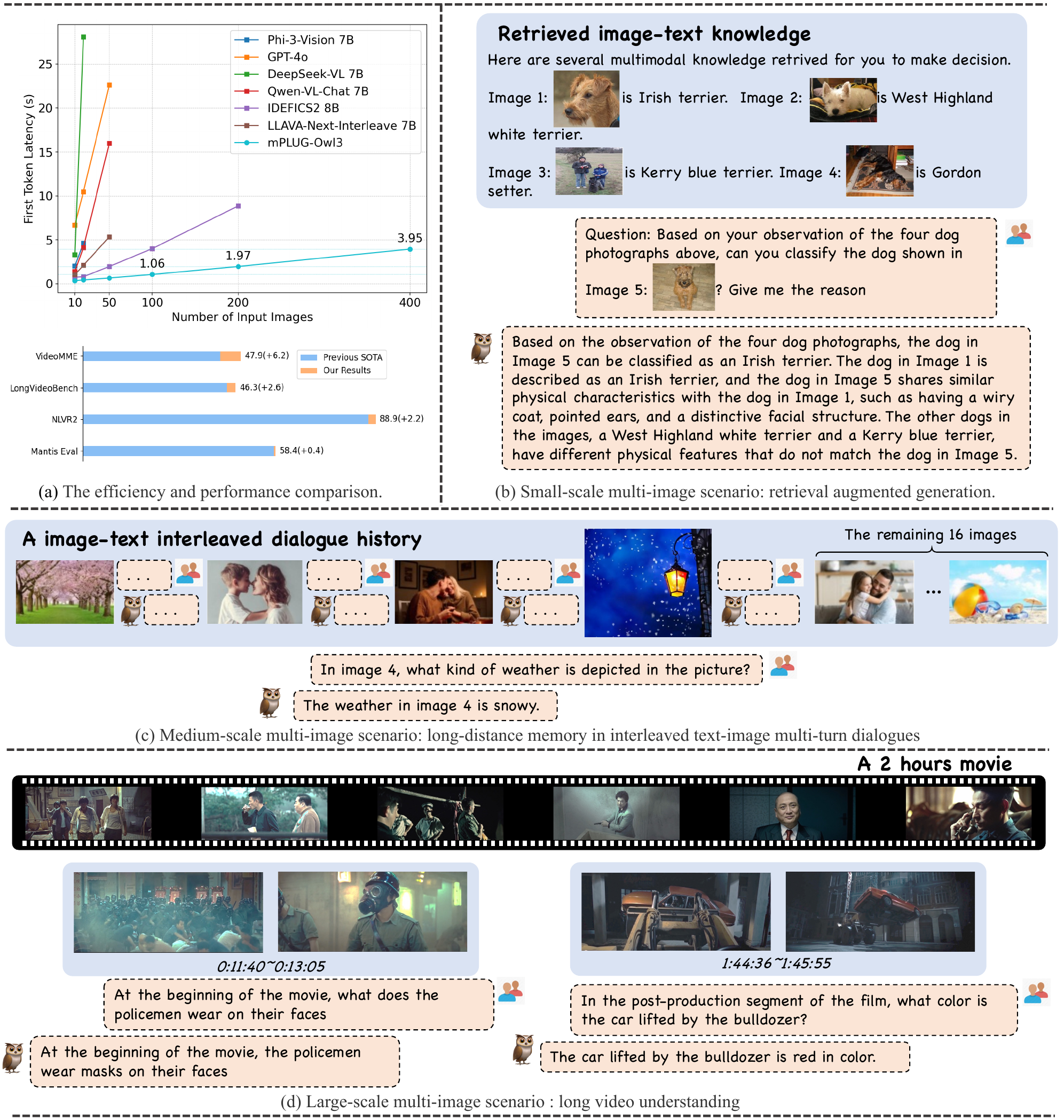}
    \caption{(a) \modelname demonstrates leading performance on video and multi-image understanding. (b,c,d) Examples of \modelname on handling different scale of multi-image scenarios.}
    \label{fig:title}
\end{figure*}
\clearpage
\begin{abstract}
Multi-modal Large Language Models (MLLMs) have demonstrated remarkable capabilities in executing instructions for a variety of single-image tasks. Despite this progress, significant challenges remain in modeling long image sequences. In this work, we introduce the versatile multi-modal large language model, mPLUG-Owl3, which enhances the capability for long image-sequence understanding in scenarios that incorporate retrieved image-text knowledge, interleaved image-text, and lengthy videos. Specifically, we propose novel hyper attention blocks to efficiently integrate vision and language into a common language-guided semantic space, thereby facilitating the processing of extended multi-image scenarios. Extensive experimental results suggest that mPLUG-Owl3 achieves state-of-the-art performance among models with a similar size on single-image, multi-image, and video benchmarks. Moreover, we propose a challenging long visual sequence evaluation named Distractor Resistance to assess the ability of models to maintain focus amidst distractions. Finally, with the proposed architecture, mPLUG-Owl3 demonstrates outstanding performance on ultra-long visual sequence inputs. We hope that mPLUG-Owl3 can contribute to the development of more efficient and powerful multimodal large language models.
\end{abstract}

\section{Introduction}
Recently, Multimodal Large Languages Models (MLLMs) ~\citep{Liu2023Llava,ye2023mplugowl,liu2024improved,ye2024mplug,chen2024internvl} have achieved rapid advancements, demonstrating strong single-image understanding capabilities. The current approaches primarily rely on vast amounts of image and text data to align Large Language Models (LLMs) ~\citep{zheng2023vicuna,Touvron2023LLaMA,Touvron2023Llama2} with visual encoders, thereby extending multimodal capabilities.

More advanced image-sequence understanding capabilities are required in practical applications, such as Multi-Image Reasoning~\citep{Suhr2018ACF,lu2021iconqa,jiang2024mantis}, Multimodal RAG~\citep{chen2022murag,lin2024fine}, Video Understanding~\citep{Xiao2021NExTQANP,Li2023MVBenchAC,fu2024video,wu2024longvideobench}, Multi-modal Agents~\citep{wang2024mobile,zhang2024ufo}, and Multi-Doc QA~\citep{tito2023hierarchical,van2023document}. The existing methods are primarily based on interleaved image-text web data for pre-training \citep{laurencon2023idefics,laurenccon2024matters} to extend multi-image capabilities or focused on the in-context abilities \citep{alayrac2022flamingo,awadalla2023openflamingo,zhao2023mmicl} within multi-image scenarios. However, these methods have not explored the in-depth comprehension or the efficiency of multi-images sufficiently, which makes it hard to support long image sequences.

%
For example, LLAVA-Next-Interleave~\citep{li2024llava} and Mantis~\citep{jiang2024mantis} directly insert visual features into textual sequences. As shown in \Cref{fig:title}, the inference latency and memory usage is dramatically increase. Flamingo~\citep{alayrac2022flamingo} simply uses a Perceiver and cross-attention layers to reduce computational overhead. This results in the loss of visual fine-grained information and leads to poor performance in both single and multi-image scenarios.

To address this challenge, we introduce mPLUG-Owl3, a new general-purpose multi-modal foundation model. mPLUG-Owl3 is designed to handle long image sequences both effectively and efficiently. mPLUG-Owl3 integrates innovative hyper attention blocks in the language model to achieve efficient interleaved vision-language semantic alignment.
%
%
Specifically, Hyper Attention introduces cross-attention parallel to the self-attention in the transformer block. The language query is reused to select and extract visual features from a lengthy visual sequence, allowing for adaptively obtaining complementary visual information that the language model lacks, based on textual semantics.

We evaluate mPLUG-Owl3 with a total of twenty benchmarks, which include single-image, multi-image, and video. Specifically, experiments encompass five visual question answering tasks, five multimodal large language model tasks, four video tasks, and six multi-image tasks. Among models of the same size, mPLUG-Owl3 achieves state-of-the-art results in 14 out of 20 benchmarks. Besides existing benchmarks, we also propose a challenging long visual sequence evaluation named Distractor Resistance. It is designed to assess the ability of models to maintain focus amidst distractions. We can observe that mPLUG-Owl3 demonstrates outstanding performance in handling ultra-long visual sequence inputs while also maintaining extremely high execution efficiency. The superior performance of the new architecture in mPLUG-Owl3 implies a trend for future multimodal large language models.

\section{\modelname}

\begin{figure*}[!ht]
    \centering
    \includegraphics[width=\textwidth]{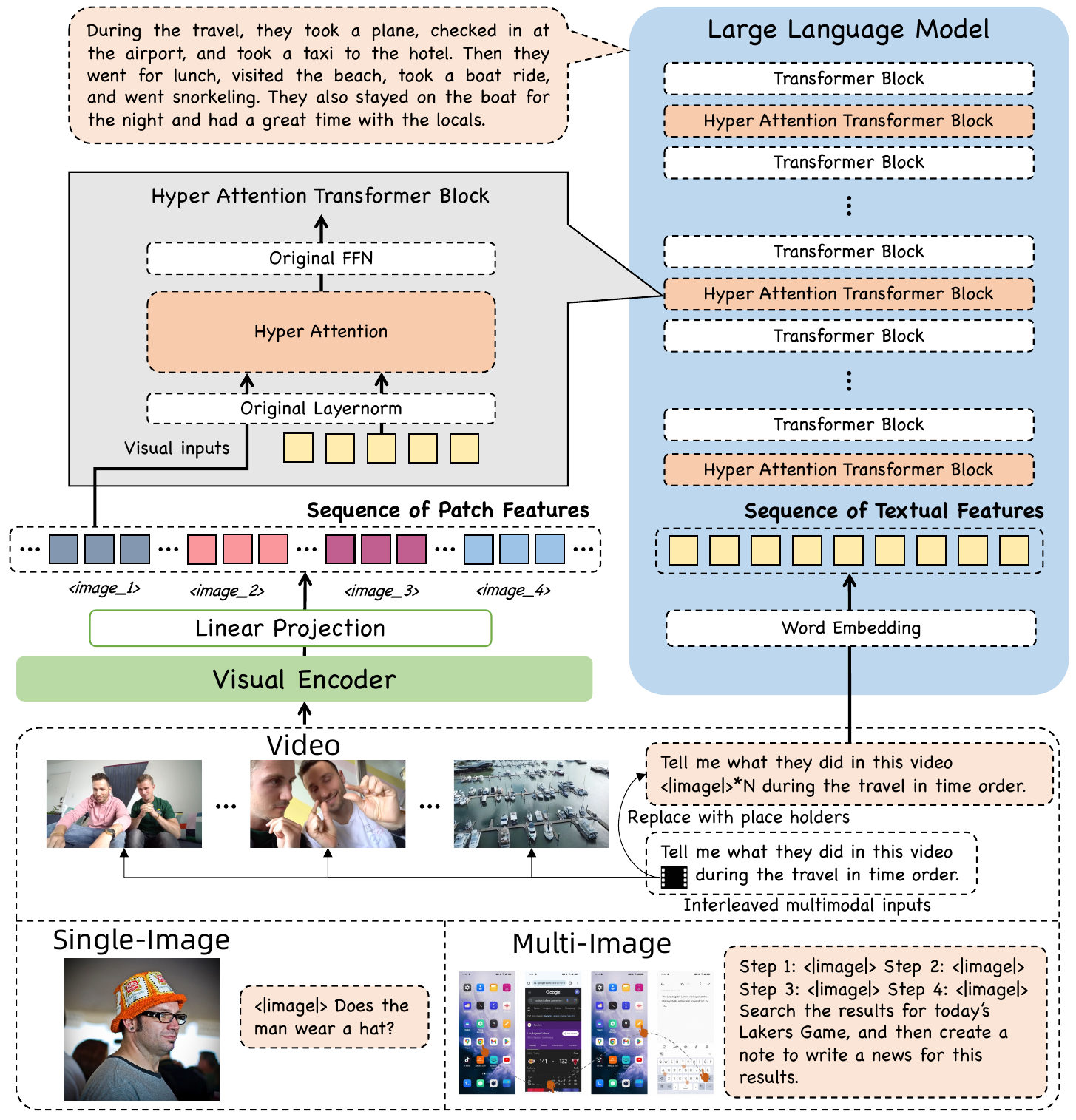}
    \caption{An overview of \modelname.}
    \label{fig:model_arch}
\end{figure*}

As illustrated in \Cref{fig:model_arch}, \modelname comprises a visual encoder, a linear projection layer, and a decoder-only language model. This architecture is commonly employed in recently proposed Multi-modal Large Language Models. Unless specified otherwise, we use Siglip-400m \citep{zhai2023sigmoid} as the visual encoder and Qwen2 \citep{yang2024qwen2} as the language model. First, we provide detailed information about our efficient architecture and its handling of various lengths of visual inputs in \Cref{sec:arch}. Additionally, we introduce the Hyper Attention module in \Cref{sec:hatb}. It is a lightweight extension designed to enhance the transformer blocks of the language model by enabling cross-attention capabilities for adaptive visual sequence utilization.

\subsection{Cross-Attention based Architecture}
\label{sec:arch}

Popular MLLMs (e.g., LLAVA-Interleave~\citep{li2024llava}, InternVL~\citep{chen2024internvl}) insert visual features into the sequence of embeddings, which can easily exhaust the language model's context window, resulting in significant memory and computational overhead. This kind of disadvantage hinders these MLLMs to modeling the long vision input such as multiple images, videos and multiple pieces high-resolution images. Moreover, visual details can be lost, when going through the language model.

Therefore, \modelname consider use cross-attention for feeding the visual information into the language model. Specifically, given a interleaved multimodal input $S=[T_1,I_1,T_2,I_2,T_3]$ (the format can be adapted to various text-image organizational structures), \modelname first extract visual features of the input images and use a linear projection to align the dimensions of visual features to be the same of the language model. The projected visual features are denoted by $\mathbf{H_{img}}=[I^t_1, I^t_2]\in \mathbb{R}^{L\times D_t}$. The text sequence are $S_{text}=[T_1,T_{img},T_2,T_{img},T_3]\in \mathbb{R}^{L\times D_t}$, where $T_{img}$ is a plain text \textit{<|image|>} to indicate the original place of the image. We feed the sequence into the word embedding to obtain text feature $\mathbf{H_{text}}$. 

In the language model, we fuse the visual features $\mathbf{H_{img}}$ into the text features $\mathbf{H^i_{text}}\in \mathbb{R}^{L\times D_t}$ of the $i^{th}$ layer through cross-attention operator. Different from Flamingo~\citep{alayrac2022flamingo} and EVLM~\citep{chen2024evlm} that insert an additional layer into each layer of transformer layer, we sparsely extend a small number of transformer blocks in the network to perform cross attention parallel with self-attention. We name the Hyper Attention Transformer Block (HATB). We discuss the design of HATB in detail in \Cref{sec:hatb}. HABT can significantly reduces the number of additional training parameters and facilitates model convergence. Besides, we observe that having fewer HATBs does not degrade the model's performance; instead, it offers the advantages of low memory consumption and high inference efficiency during inference. For a language model consisting of $N$ layers, we start from layer 0 and uniformly extend $K$ layers to HATB. Specifically, for Qwen2, we select layers $[0, 9, 17, 25]$.

\subsection{Hyper Attention Transformer Block}
\label{sec:hatb}


In this section, we specifically introduce the Hyper Attention Transformer Block used in \modelname. The cross-attention structure employed in Flamingo, as shown in \Cref{fig:habt} (a), has been widely utilized in constructing MLLMs (e.g., IDEFICS~\citep{laurencon2023idefics}, EVLM~\citep{chen2024evlm}). However, this structure presents three main drawbacks: it introduces a large number of additional parameters, which results in significant memory and computational overhead; the knowledge learned by the language model cannot benefit the understanding of visual inputs; the cross attention does not fully take into account the original positions of images in the interleaved sequence, which limits the performance of these models in multi-image scenarios. In response to these issues, we propose a lightweight Hyper Attention Transformer Block, illustrated in \Cref{fig:habt} (b). This block introduces a small number of parameters and extends self-attention capabilities to perform both intra-text self-attention and inter-modal cross-attention between text and images in parallel. It also introduces a Multimodal-Interleaved Rotary Position Embedding (MI-Rope) to maintain the position information of images. The extended modifications are detailed below:

\begin{figure*}[!ht]
    \centering
    \includegraphics[width=\textwidth]{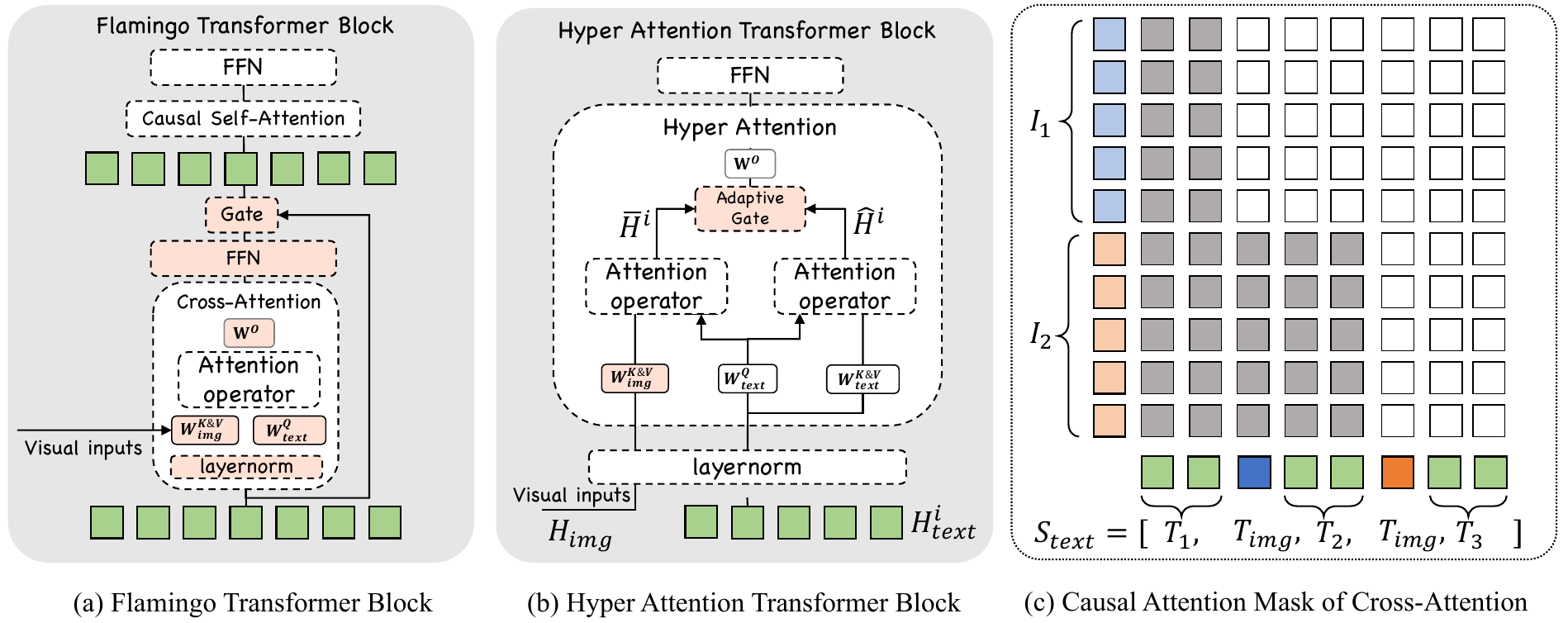}
    \caption{Comparison between Flamingo Transformer Block (a) and Hyper Attention Transformer Block (b). Pink indicates that the module is additionally introduced. (c) presents the causal attention mask strategy of cross attention in Hyper Attention in a image-text interleaved scenario. The gray block denotes the attention score is ignored. $T_{img}$ denotes the token of plain text \textit{<|image|>}.}
    \label{fig:habt}
\end{figure*}

\paragraph{Shared Input Layernorm.} The visual feature $\mathbf{H_{img}}$ and the $i^{th}$ layer's text features $\mathbf{H^i_{text}}$, although sharing the same dimensionality, originate from different distributions. Hence, both sets of features are initially normalized using a LayerNorm module. Our findings indicate that employing the LayerNorm module already integrated within the transformer block results in better convergence compared to training a separate layer normalization module specifically for the visual features. This improvement is attributed to the compatibility of the mean and variance of the outputs from the integrated LayerNorm module with the distribution characteristics of the pre-trained language model.

\paragraph{Modality-Specific Key-Value Projection.} In cross-attention, the \textbf{Query} is derived from textual data, while the \textbf{Key} and \textbf{Value} are extracted from visual features. Inspired by \citet{ye2024mplug}, we construct a weight matrix $\mathbf{W}^{K\&V}_{img} \in \mathbb{R}^{2D \times D}$ to generate the \textbf{Key} and \textbf{Value} for the visual features. This matrix is initialized using the weights from the language model’s KV (Key-Value) projection. Furthermore, the query vector from the self-attention mechanism is repurposed as the \textbf{Query} in the cross-attention. The computation procedure for self-attention remains unchanged. This design is beneficial as it preserves more specific visual information and allows for the adaptive supplementation of visual information that the language model lacks, based on textual semantics.

\paragraph{Visual Position Modeling in Attention.} For models that process multiple images, positional encoding is essential to correctly understanding interleaved image-text input. Existing cross-attention models, such as Flamingo~\citep{alayrac2022flamingo} and IDEFICS~\citep{laurencon2023idefics}, do not assign position embeddings to visual inputs, leading to suboptimal performance in scenarios involving multiple images. To accurately represent the original positions of images in interleaved sequences, we develope a Multimodal-Interleaved Rotary Position Embedding, which we name MI-Rope. Specifically, for each visual feature $I_n$ of image $n$, we pre-record the position index of its placeholder $T_{img}$ in the interleaved sequence $S_{text}$. All patches of $I_n$ share $T_n$'s positional encoding to obtain the rotary embedding. This ensures that the positional encoding of the image not only reflects the order among images but also reveals its position in the textual context. We also use a causal attention mask in cross attention. As shown in \Cref{fig:habt} (c), for a text sequence $S=[T_1,T_{img},T_2,T_{img},T_3]$, each text token can only attend the visual features that precede it. Then, HATB simultaneously performs cross-attention and self-attention, denoting the resulting hidden states as $\mathbf{\bar{H}^i}$ and $\mathbf{\hat{H}^i}$.

\paragraph{Adaptive Gating} Existing implementations of cross-attention utilize a learnable scale to regulate the extent of information transfer from the image to the language model. However, the semantics of language are ignored. Consequently, we introduce an adaptive gate that obtains the gate value based on the textual features:

\begin{align}
    \mathbf{g} &= \mathrm{Sigmoid}(\mathbf{W}^T_{gate}\mathbf{\hat{H}^i}) \\
    \mathbf{H^i_{fused}} &= \mathbf{\bar{H}^i_{text}} * \mathbf{g} + \mathbf{\hat{H}^i_{text}} * (1-\mathbf{g})
\end{align}

The $\mathbf{H^i_{fused}}$ is passed to the FFN and fed to the next layer of transformer.


\section{Implement Details}
\subsection{Training Paradigm}
We adopt a three-stage training approach for \modelname. Initially, we pre-train \modelname using image-text pairs to achieve robust multimodal alignment. In the second stage, we leverage diverse datasets that include image and video captions to enhance the model's ability to understand multiple images. Finally, we fine-tune \modelname using a mixture of supervised data, encompassing tasks involving both single and multiple images, to ensure comprehensive performance. The statistics of the datasets we used are presented in \Cref{tab:dataset}, and the training settings are detailed in \Cref{tab:training}.

\begin{table}[h!]
  \centering
  \resizebox{1\linewidth}{!}{
  \begin{tabular}{@{}ll|ll|ll@{}}
    \toprule
    \multicolumn{2}{c}{\textbf{Stage 1: Pretraining}} & \multicolumn{2}{c}{\textbf{Stage 2: Multi-Image Training}}& \multicolumn{2}{c}{\textbf{Stage 3: Self-Supervised Fintuning}} \\
    \textbf{Dataset Name} & \textbf{Percentage} & \textbf{Dataset Name} & \textbf{Percentage} & \textbf{Dataset Name} & \textbf{Percentage} \\ \midrule
    DataComp-1B           & 35.22\%             & ShareGPTVideo         & 34.63\%             & LLAVA-SFT             & 57.95\%             \\
    LAION-en              & 26.07\%             & Selective Caption       & 19.29\%             & The Cauldron          & 12.50\%             \\
    COYO-700M             & 14.47\%             & LLAVA-Interleave      & 16.69\%             & Mantis                & 10.41\%             \\
    COYO-700M-OCR          & 9.60\%              & VATEX                 & 15.77\%             & LLAVA-Interleave      & 9.26\%              \\
    LAION-zh              & 7.73\%              & Text Reading          & 7.36\%              & ALLAVA                & 6.95\%              \\
    Wukong                & 5.64\%              & Interleaved Caption   & 5.25\%              & ShareGPTVideo-QA 240K & 2.02\%              \\
    CC12M                 & 0.81\%              & MMDU           &  1.01\%            & Video Instruct 100K   & 0.84\%              \\
    Others                  & 0.46\%              &         -        &    -          & MSRTT/MSVD Caption    & 0.06\%              \\
    \bottomrule
  \end{tabular}
  }
  \caption{Dataset percentages used in Pretraining, Multi-Image Training, and Self-Supervised Fintuning. Others include CC3M, OCR-CC, COCO and SBU.}
  \label{tab:dataset}
\end{table}
\begin{table}[htbp]
\resizebox{1\linewidth}{!}{
\begin{tabular}{@{}ll|ccc@{}}
\toprule
\multicolumn{2}{l|}{\textbf{Setting}}                    & \textbf{Stage 1: Pretraining}                                                                    & \textbf{Stage 2: Multi-Image Training}                                          & \textbf{Stage 3: Self-Supervised Fintuning}                                     \\ \midrule
\multirow{5}{*}{Training}     & Learning Rate (Max, Min) & (1e-3, 1e-5)                                                                                     & (2e-5, 1e-7)                                                                    & (2e-5, 1e-7)                                                                    \\ \cmidrule(l){2-5} 
                              & Global Batch Size        & 2048                                                                                             & 1024                                                                            & 1024                                                                            \\ \cmidrule(l){2-5} 
                              & Training Steps           & 20K                                                                                              & 3K                                                                              & 11K                                                                             \\ \cmidrule(l){2-5} 
                              & Warmup ratio             & \multicolumn{3}{c}{0.03}                                                                                                                                                                                                                                             \\ \cmidrule(l){2-5} 
                              & Trainable Modules        & \begin{tabular}[c]{@{}c@{}}Linear Projection\\ Visual KV Projection\\ Adaptive Gate\end{tabular} & \begin{tabular}[c]{@{}c@{}}Linear Projection\\ Full Language Model\end{tabular} & \begin{tabular}[c]{@{}c@{}}Linear Projection\\ Full Language Model\end{tabular} \\ \midrule
\multirow{2}{*}{Model}        & Resolution               & $384^{2}$                                                                                              & up to $384^{2} \times 6$                                                                             & up to $384^{2} \times 6$                                                                               \\ \cmidrule(l){2-5} 
                              & Sequence Length          & 768                                                                                              & 4096                                                                            & 4096                                                                            \\ \midrule
\multirow{4}{*}{Accelerating} & Precision                & \multicolumn{3}{c}{Mixed-precision FP16/BF16}                                                                                                                                                                                                                        \\ \cmidrule(l){2-5} 
                              & ZeRO Optimization        & \multicolumn{3}{c}{Zero-1}                                                                                                                                                                                                                                           \\ \cmidrule(l){2-5} 
                              & Gradient Checkpointing   & No.                                                                                              & Yes.                                                                            & Yes.                                                                            \\ \cmidrule(l){2-5} 
                              & Model Parallel           & TP=1                                                                                             & TP=4                                                                            & TP=4                                                                            \\ \bottomrule
\end{tabular}
}

  \caption{The training settings across three stages: Pretraining, Multi-Image Training, and Self-Supervised Finetuning.}
  \label{tab:training}
\end{table}
\subsubsection{Pre-training}
We collect image-text pairs from public datasets, including Conceptual Captions (CC3M/CC12M) \citep{changpinyo2021cc3m12m}, COCO \citep{lin2014coco}, Laion \citep{schuhmann2022laion}, COYO \citep{kakaobrain2022coyo-700m}, DataComp \citep{gadre2023datacomp}, Wukong~\citep{gu2022wukong}, ImageNet~\citep{deng2009imagenet}, OCR-CC~\citep{yang2021tap} and SBU~\citep{ordonez2011im2text}. We randomly sample a subset consists of 41 million image-text pairs for pre-training. During pre-training, only the newly introduced modules are trainable, which include the linear layer following the vision encoder, the visual KV projection, and the Adaptive Gate in the Hyper Attention Transformer Block.

\subsubsection{Multi-image Pre-training}
In the multi-image pre-training stage, we collected three types of data to enhance the model's multi-image understanding capabilities:

\begin{itemize}

\item Interleaved data: We utilize sources such as MMDU~\citep{liu2024mmdu} and LLaVA-Interleave~\citep{li2024llava} for multi-image data. Additionally, from LLaVA-Recap 558K, we randomly sample 3 to 6 images and combine their image-caption pairs into an interleaved format to create Interleaved Captions. We also consider sampling 4 images and requiring a description of one among them to form Selective Captions.

\item Text-rich data: We use text reading and key point generation data proposed by UReader~\citep{ye2023ureader}, enabling the model to reconstruct the text contained within text-rich images and TO extract key points. These data help the model learn the original text structure from the pieces of high-resolution images.

\item Video data: We adopt annotated data from ShareGPTVideo~\citep{zhang2024direct}, which includes 900K caption entries and 240K question-answering instances. We also incorporate Chinese and English video caption data from VATEX~\citep{wang2019vatex}. For training with video data, we consistently sample 8 frames per video. 

\end{itemize}

Both linear projection and the full language model are trainable. With the help of tensor parallelism, the model is spilted into four parts, effectively reducing the memory usage on a single GPU to between 32 and 40 GB.

\subsubsection{Supervised-Finetuning}
In Supervised-Finetuning stage, \modelname is trained with an extensive and diverse assembly of instruction tuning datasets aimed at enhancing its instruction-following capability. The datasets include LLaVA-SFT-665K~\citep{liu2024improved}, The Cauldron~\citep{laurenccon2024matters}, Mantis~\citep{jiang2024mantis}, LLaVA-Interleave~\citep{li2024llava}, ALLaVA~\citep{chen2024allava}, ShareGPTVideo-QA 240K~\citep{zhang2024direct}, Video Instruct 100K~\citep{Maaz2023VideoChatGPT}, MSR-VTT~\citep{xu2016msr} and MSVD Caption~\citep{chen2011collecting}. We keep the same training setting as the Multi-image Pre-training stage.

\subsection{High-resolution Image Processing}
Inspired by UReader~\citep{ye2023ureader}, we introduce a similar adaptive method for image cropping. For a given image, we select from the cropping grids (2,2), (1,3), (1,4), (3,1), (4,1), (2,3), and (3,2) that most closely matches the shape of the input image. Additionally, we retain a global version of the original image. During the Supervised-Finetuning stage, for datasets rich in text, we perform cropping with a probability of 100\%. For datasets containing a single image without text, we apply cropping with a probability of 20\%. For datasets containing multiple images or videos, we do not perform cropping. During evaluation, cropping is enabled only for single-image tasks.

\subsection{Video Processing}
For videos, we sample 8 frames per video by default. Meanwhile, we replace the video markers in the text with multiple \textit{<|image|>} placeholders corresponding to the number of sampled frames.

\section{Experiments}

\begin{table}[btp]
\centering 
\resizebox{0.8\linewidth}{!}{
\begin{tabular}{@{}lcccccc@{}}
\toprule
\multicolumn{1}{l|}{Model}         & \multicolumn{1}{c|}{\# Param} & VQAv2         & OK-VQA        & GQA           & VizWizQA      & TextVQA       \\ \midrule
\multicolumn{1}{l|}{CogVLM}        & \multicolumn{1}{c|}{17B}      & 82.3          & 64.8          & -             & -             & 70.4          \\
\multicolumn{1}{l|}{EVLM-Chat}     & \multicolumn{1}{c|}{32B}      & 81.9          & -             & 64.4          & 47.3          & 67.5          \\
\multicolumn{1}{l|}{Flamingo}      & \multicolumn{1}{c|}{80B}      & 81.3          & 50.6          & -             & 57.2          & 54.7          \\ \midrule
\multicolumn{7}{l}{8B-level MLMMs}                                                                                                                 \\ \midrule
\multicolumn{1}{l|}{Qwen-VL-Chat}  & \multicolumn{1}{c|}{9B}       & 78.2          & 56.6          & 57.5          & 38.9          & 63.8          \\
\multicolumn{1}{l|}{Idefics1}      & \multicolumn{1}{c|}{9B}       & 68.8          & 50.4          & -             & -             & 39.3          \\
\multicolumn{1}{l|}{Flamingo}      & \multicolumn{1}{c|}{9B}       & 51.8          & 44.7          & -             & -             & 46.3          \\
\multicolumn{1}{l|}{InstructBLIP}  & \multicolumn{1}{c|}{7B}       & 75.2          & 45.2          & 49.2          & 34.5          & 33.6          \\
\multicolumn{1}{l|}{mPLUG-Owl2}    & \multicolumn{1}{c|}{8B}       & 79.4          & {\ul 57.7}    & 56.1          & 54.5          & 58.2          \\
\multicolumn{1}{l|}{LLAVA-1.5}     & \multicolumn{1}{c|}{8B}       & 78.5          & -             & 62.0          & 50.0          & 58.2          \\
\multicolumn{1}{l|}{LLAVA-Next}    & \multicolumn{1}{c|}{8B}       & {\ul 81.8}    & -             & {\ul 64.2}    & 57.6          & 64.9          \\
\multicolumn{1}{l|}{VILA-1.5}      & \multicolumn{1}{c|}{8B}       & 80.9          & -             & 61.9          & {\ul 58.7}    & 66.3          \\
\multicolumn{1}{l|}{Idefics2}      & \multicolumn{1}{c|}{8B}       & 80.8          & 53.5          & -             & -             & \textbf{70.4} \\
\multicolumn{1}{l|}{Mantis-SigLIP} & \multicolumn{1}{c|}{8B}       & 74.9          & 55.4          & -             & -             & 59.2          \\ \midrule
\multicolumn{1}{l|}{mPLUG-Owl3}    & \multicolumn{1}{c|}{8B}       & \textbf{82.1} & \textbf{60.1} & \textbf{65.0} & \textbf{63.5} & {\ul 69.0}    \\ \bottomrule
\end{tabular}
}
\caption{\textbf{Performance comparison on visual question answering.} The accuracy is reported. We use \textbf{bold} to mark the highest score and {\ul underline} to mark the second highest of 8B-level MLLMs.}
\label{tab:vqa}
\end{table}
\subsection{Visual Question Answering Benchmarks}

We conduct experiments on a diverse set of visual question answering benchmarks, including VQAv2~\citep{Goyal2016MakingTV}, OK-VQA~\citep{marino2019okvqa}, GQA~\citep{hudson2019gqa}, VizWizQA~\citep{bigham2010vizwiz}, and TextVQA~\citep{singh2019towards}. The VQAv2 dataset is currently the largest visual question answering dataset available. OK-VQA involves questions that require external knowledge beyond multimodal inputs. GQA is designed to validate the model's reasoning capabilities. VizWizQA is constructed from question-answer pairs sourced from visually impaired users. TextVQA focuses more on evaluating the model's ability to understand text in natural scenes. These datasets are strategically selected to thoroughly evaluate our model's ability to understand and reason across various visual contexts and knowledge domains. 
\Cref{tab:vqa} presents the comparison results between \modelname and State-of-the-Art multimodal large language models, including CogVLM~\citep{Wang2023CogVLMVE}, EVLM-Chat~\citep{chen2024evlm}, flamingo~\citep{alayrac2022flamingo}, Qwen-VL-Chat~\citep{Bai2023QwenVL}, Idefics~\citep{laurencon2023idefics}, InstructBLIP~\citep{Dai2023InstructBLIP}, mPLUG-Owl2~\citep{ye2024mplug}, LLaVA-1.5~\citep{liu2024improved}, LLaVA-Next~\citep{liu2024llavanext}, VILA-1.5~\citep{lin2023vila}, Idefics2~\citep{laurenccon2024matters}, Mantis-SigLIP~\citep{jiang2024mantis}.


\modelname outperforms 8B-level language models in VQAv2, OK-VQA, GQA, and VizWizQA. Furthermore, it surpasses the 32B-parameter EVLM\footnote{EVLM does not provide the number of parameters for its cross module. The parameter count in this table is estimated based on its model architecture.} in GQA and VizWizQA. In TextVQA, although \modelname's performance is slightly lower than that of Idefics2, it still exceeds that of other 8B models. It is noteworthy that, despite having 8B parameters, \modelname exhibits superior inference speed and memory efficiency compared to models of the same scale, thanks to the introduction of Hyper Attention.

\subsection{General MLLM Benchmarks}

We evaluate \modelname on various single-image general multimodal large language model benchmarks including MMBench-EN/CN~\citep{liu2023mmbench}, MM-Vet~\citep{yu2023mmvet}, POPE~\citep{Li2023pope} and AI2D~\citep{kembhavi2016diagram}. MMBench provides a comprehensive evaluation of a model's multimodal capabilities in both Chinese and English contexts. MM-Vet assesses the multimodal conversational abilities of a model using GPT-4 evaluation. POPE can evaluate the extent of multimodal hallucinations in a model. AI2D assesses a model's ability to understand science diagrams inputs.

\Cref{table:mllm} shows that \modelname achieves state-of-the-art performance on MMBench-EN, MMBench-CN, MM-Vet and POPE across 8B-level models such as OpenFlamingo~\citep{awadalla2023openflamingo}, Cambrian~\citep{tong2024cambrian} and MiniCPM-Llama3-V2.5~\citep{yao2024minicpmvgpt4vlevelmllm}.
It also matches or surpasses the performance of larger models such as CogVLM~\citep{Wang2023CogVLMVE} and EVLM-Chat~\citep{chen2024evlm}. \modelname does not achieve state-of-the-art performance on the AI2D dataset. Due to limited training resources, we do not fine-tune the vision encoder, which restricts its performance in scenarios rich in text.

\begin{table}[btp]
\centering 
\resizebox{0.8\linewidth}{!}{
\begin{tabular}{@{}lcccccc@{}}
\toprule
\multicolumn{1}{l|}{Model}               & \multicolumn{1}{c|}{\# Param} & MMB-EN        & MMB-CN        & MM-Vet        & POPE          & AI2D          \\ \midrule
\multicolumn{1}{l|}{CogVLM}              & \multicolumn{1}{c|}{17B}      & 65.8          & 69.8          & 52.8          & 88.0          & 63.3          \\
\multicolumn{1}{l|}{EVLM-Chat}           & \multicolumn{1}{c|}{32B}      & 76.9          & 76.9          & -             & 89.7          & 76.0          \\
\multicolumn{1}{l|}{InstructBLIP}        & \multicolumn{1}{c|}{13B}      & 38.3          & -             &               & 81.5          & -             \\ \midrule
\multicolumn{7}{l}{8B-level MLMMs}                                                                                                                       \\ \midrule
\multicolumn{1}{l|}{LLAVA-1.5}           & \multicolumn{1}{c|}{8B}       & 64.3          & 58.3          & 31.1          & 85.9          & 55.5          \\
\multicolumn{1}{l|}{OpenFlamingo}        & \multicolumn{1}{c|}{9B}       & 32.4          & 14.4          & 24.8          & -             & 31.7          \\
\multicolumn{1}{l|}{mPLUG-Owl2}          & \multicolumn{1}{c|}{8B}       & 64.5          & -             & 36.2          & -             & 55.7          \\
\multicolumn{1}{l|}{LLAVA-Next}          & \multicolumn{1}{c|}{8B}       & 67.4          & 60.6          & \textbf{43.9} & 86.5          & 66.6          \\
\multicolumn{1}{l|}{LLAVA-Interleave}    & \multicolumn{1}{c|}{8B}       & -             & -             & -             & {\ul 86.8}    & 73.9          \\
\multicolumn{1}{l|}{VILA1.5}             & \multicolumn{1}{c|}{8B}       & 72.3          & 66.2          & 38.3          & 84.4          & -             \\
\multicolumn{1}{l|}{Idefics2}            & \multicolumn{1}{c|}{8B}       & {\ul 75.7}    & 68.6          & 34.0          & 86.2          & 72.3          \\
\multicolumn{1}{l|}{Cambrian}            & \multicolumn{1}{c|}{8B}       & 74.6          & 67.9          & -             & -             & {\ul 74.6}    \\
\multicolumn{1}{l|}{MiniCPM-Llama3-V2.5} & \multicolumn{1}{c|}{8B}       & \textbf{77.6} & {\ul 73.8}    & -             & -             & \textbf{78.4} \\
\multicolumn{1}{l|}{Mantis-SigLIP}       & \multicolumn{1}{c|}{8B}       & 68.7          & -             & -             & -             & -             \\ \midrule
\multicolumn{1}{l|}{mPLUG-Owl3}          & \multicolumn{1}{c|}{8B}       & \textbf{77.6} & \textbf{74.3} & {\ul 40.1}    & \textbf{88.2} & 73.4          \\ \bottomrule
\end{tabular}
}
    \caption{\textbf{Zero-shot multi-modal evaluation on multi-modal benchmarks.} The overall scores are reported for evaluation. We use \textbf{bold} to mark the highest score and {\ul underline} to mark the second highest of 8B-level MLLMs.}
    \label{table:mllm}
\end{table}

\subsection{Multi-image and Video Benchmark}
We also evaluate the performance of mPLUG-Owl3 on video and multi-image benchmarks, as it is capable of processing multiple images with an interleaved format. we include VideoChat2~\citep{Li2023MVBenchAC}, Video-LLaMA2~\citep{cheng2024videollama}, Video-ChatGPT~\citep{Maaz2023VideoChatGPT}, ShareGPT4Video~\citep{chen2024sharegpt4video}, PLLaVA~\citep{xu2024pllava}, Idefics2~\citep{Laurenccon2024WhatMW}, Mantis-SigLIP~\citep{jiang2024mantis} and LLAVA-Interleave~\citep{li2024llava}.

The results of video evaluation is shown in \Cref{tab:video}. The NextQA~\citep{Xiao2021NExTQANP} and MVBench~\citep{Li2023MVBenchAC} are short video benchmarks, with video durations all less than one minute. \modelname achieves performance comparable to state-of-the-art models. For benchmarks like VideoMME~\citep{fu2024video} and LongVideoBench~\citep{wu2024longvideobench}, with longer video durations up to one hour, \modelname significantly outperforms existing models. It demonstrates that \modelname is highly suitable for understanding videos with various durations.

\begin{table}[tbp]
\resizebox{1\linewidth}{!}{
\begin{tabular}{@{}l|c|cccc@{}}
\toprule
Model            & \# Param & NextQA        & MVBench       & VideoMME w/o sub & LongVideoBench-val \\ \midrule
VideoChat2       & 8B       & 68.6          & 51.9          & 43.8             & 36.0               \\
Video-LLaMA2     & 8B       & -             & \textbf{54.6} & {\ul 47.9}       & -                  \\
Video-ChatGPT    & 8B       & -             & 32.7          & -                & -                  \\
ShareGPT4Video   & 8B       & -             & -             & 39.9             & 39.7               \\
PLLaVA           & 8B       & -             & 46.6          & -                & 40.2               \\
Idefics2         & 8B       & -             & 29.7          & -                & {\ul 49.7}         \\
Mantis-SigLIP    & 8B       & -             & 50.2          & -                & 47.0               \\
LLAVA-Interleave & 8B       & {\ul 78.2}    & 53.1          & -                & -                  \\ \midrule
mPLUG-Owl3       & 8B       & \textbf{78.6} & {\ul 54.5}    & \textbf{53.5}    & \textbf{52.1}      \\ \bottomrule
\end{tabular}
}
\caption{\textbf{Multi-modal evaluation on video understanding benchmarks.} The overall scores are reported for evaluation. We use \textbf{bold} to mark the highest score and {\ul underline} to mark the second highest.}
\label{tab:video}
\end{table}

\Cref{tab:multi_image} presents the the evaluation results on multi-image understanding. NLVR2~\citep{Suhr2018ACF} and Mantis-Eval~\citep{jiang2024mantis} test the model's ability to perform logical reasoning based on the content of multiple images. MathVerse-mv~\citep{li2024llava} and SciVerse-mv~\citep{li2024llava} evaluate the model's multi-image mathematical and scientific capabilities. We use the version released by llava-next-interleave for comparison with its reported results. BLINK~\citep{Fu2024BLINKML} and Q-Bench2~\citep{zhang2024benchmark} test the model's multi-image question answering ability based on low-level visual perception. We compare \modelname with models support image-text interleaved inputs such as Qwen-VL-Chat~\citep{Bai2023QwenVL}, InstructBLIP~\citep{Dai2023InstructBLIP}, CogVLM~\citep{Wang2023CogVLMVE}, VideoLLaVA~\citep{Lin2023VideoLLaVALU}, VILA~\citep{lin2023vila}, Idefics2~\citep{Laurenccon2024WhatMW}, Mantis-SigLIP~\citep{jiang2024mantis} and LLAVA-Interleave~\citep{li2024llava}.

\modelname surpasses existing models in both NLVR2 and Mantis-Eval. On MathVerse-mv and SciVerse-mv, it can be observed that \modelname significantly outperforms LLaVA-Interleave. However, on BLINK, \modelname performs weaker than LLaVA-Interleave. We note that this dataset requires models to possess low-level visual perception capabilities for fine details in images, and \modelname's ability may be limited due to the vision encoder being frozen during training. On the Q-Bench2, which evaluates a model's ability to discern low-level visual differences across multiple images globally, \modelname achieves performance comparable to the state-of-the-art.

\begin{table}[htp]
\resizebox{1\linewidth}{!}{
\begin{tabular}{@{}l|c|cccccc@{}}
\toprule
\textbf{Model}   & \# Param & NLVR2         & Mantis-Eval   & MathVerse-mv  & SciVerse-mv   & BLINK         & Q-Bench2       \\ \midrule
Qwen-VL-Chat     & 8B       & 58.7          & 39.2          & -             & -             & 31.2          & 45.9          \\
InstructBLIP     & 8B       & 60.3          & 45.6          & -             & -             & 42.2          & 44.3          \\
CogVLM           & 17B      & 58.6          & 45.2          & -             & -             & 41.5          & 53.2          \\
VideoLLaVA       & 8B       & 56.5          & 35.9          & -             & -             & 38.9          & 45.7          \\
VILA             & 8B       & 76.5          & 51.2          & -             & -             & 39.3          & 45.7          \\
Idefics2         & 8B       & 86.9          & 48.9          & -             & -             & 45.2          & 57.0          \\
Mantis-SigLIP    & 8B       & 87.4          & 59.5          & -             & -             & 46.4          & 69.9          \\
LLAVA-Interleave & 8B       & {\ul 88.8}    & {\ul 62.7}    & {\ul 32.8}    & {\ul 31.6}    & \textbf{52.6} & \textbf{74.2} \\ \midrule
mPLUG-Owl3       & 8B       & \textbf{90.8} & \textbf{63.1} & \textbf{65.0} & \textbf{86.2} & {\ul 50.3}    & {\ul 74.0}    \\ \bottomrule
\end{tabular}
}
\caption{\textbf{Multi-modal evaluation on multi-image understanding benchmarks.} The overall scores are reported for evaluation. We use \textbf{bold} to mark the highest score and {\ul underline} to mark the second highest.}
\label{tab:multi_image}
\end{table}

To more comprehensively investigate the fine-grained abilities of \modelname in multi-image scenarios, we conduct experiments on MI-Bench~\citep{liu2024mibench}, a recently proposed multi-image benchmark. We exclude Fine-Grained Visual Recognition from evaluation because it consists of images from mini-ImageNet that may have been seen by existing models.

\Cref{tab:mibench} shown that \modelname achieves state-of-the-art performance on aspects of General Comparison, Subtle Difference, Temporal Reasoning, Logical Reasoning and Text-Rich Images across popular open-sourced MLLMs. It also outperform GPT-4V and GPT-4o on General Comparison. The results demonstrates that our model possesses robust capabilities in various multi-image input scenarios. The Hyper Attention structure of \modelname better preserves the original visual features, enabling it to excel in single-image tasks as well. And this type of multimodal knowledge also assists it in more accurately completing multi-image tasks.

\begin{table}[tbp]
\centering
\resizebox{0.8\linewidth}{!}{
\begin{tabular}{@{}lcccccccc@{}}
\toprule
\multicolumn{1}{l|}{Model}      & GC            & SD            & VR            & TR            & \multicolumn{1}{c|}{LR}            & TRI           & VTK           & TVK           \\ \midrule
\multicolumn{9}{l}{Closed-source MLLMs}                                                                                                                                              \\ \midrule
\multicolumn{1}{l|}{GPT-4o}     & 80.7          & 90.5          & 46.8          & 68.0          & \multicolumn{1}{c|}{69.8}          & 74.8          & 54.7          & 63.3          \\
\multicolumn{1}{l|}{GPT-4V}     & 72.8          & 79.2          & 45.8          & 61.8          & \multicolumn{1}{c|}{66.3}          & 71.0          & 52.0          & 56.0          \\ \midrule
\multicolumn{9}{l}{Open-source MLLMs}                                                                                                                                                \\ \midrule
\multicolumn{1}{l|}{mPLUG-Owl2} & 64.2          & 40.1          & 35.6          & 30.7          & \multicolumn{1}{c|}{41.3}          & 39.0          & 17.0          & 25.6          \\
\multicolumn{1}{l|}{MMICL}      & 53.7          & 46.4          & \textbf{41.1} & 47.0          & \multicolumn{1}{c|}{59.6}          & 27.6          & 22.1          & 35.9          \\
\multicolumn{1}{l|}{Idefics2-I} & 83.1          & 49.7          & 32.6          & 44.8          & \multicolumn{1}{c|}{56.4}          & 43.9          & 25.6          & 39.0          \\
\multicolumn{1}{l|}{Mantis}     & 83.0          & 54.1          & 37.6          & 45.5          & \multicolumn{1}{c|}{63.4}          & 37.7          & 26.4          & 41.7          \\ \midrule
\multicolumn{1}{l|}{mPLUG-Owl3} & \textbf{86.4} & \textbf{70.1} & 33.0          & \textbf{46.8} & \multicolumn{1}{c|}{\textbf{67.2}} & \textbf{50.1} & \textbf{31.1} & \textbf{48.8} \\ \bottomrule
\end{tabular}
}
\caption{\textbf{Multi-image evaluation on MI-Bench~\citep{liu2024mibench}}. We use \textbf{bold} to mark the highest score of open-sourced multimodel large language models. The evaluation consists of the following tasks: General Comparison (GC), Subtle Difference (SD), Visual Referring (VR), Temporal Reasoning (TR), Logical Reasoning (LR), Text-Rich Images (TRI), and Vision-linked Textual Knowledge (VTK).}
\label{tab:mibench}
\end{table}

\subsection{Ablation Studies}
We adopt the training methods of LLaVA-1.5~\citep{liu2024improved} using the same datasets to conduct our ablation study. Additionally, we employ the Qwen1.5 7B as our language model. To validate the single-image understanding capabilities of our structures, we use datasets such as GQA and TextVQA (with OCR). Furthermore, we examine the generalization capabilities of our structures in multi-image understanding and video comprehension by conducting zero-shot evaluations on benchmarks including MvBench, VideoMME, NLVR2, and Mantis-Eval.

\subsubsection{Cross Attention Integration}

There are two primary methods to integrate Cross-Attention into the transformer block: one method positions it prior to the self-attention (referred to as Pre-Cross-Attention), while the other places it subsequent to the self-attention (referred to as Post-Cross-Attention). We analyze both configurations and compare them to the concatenate-based method and our novel Hyper Attention in \modelname. Specifically, for Pre-Cross-Attention, it is positioned before the layer normalization at the input stage of the Transformer block. Conversely, for Post-Cross-Attention, it is positioned after the layer normalization that follows the self-attention stage. Both attention mechanisms employ a gating mechanism to fuse the multimodal representations effectively.

\begin{table}[tbp]
\centering
\resizebox{0.9\linewidth}{!}{
\begin{tabular}{@{}l|cccccc@{}}
\toprule
Attention Structure  & GQA           & TextVQA       & MvBench       & VideoMME      & NLVR2         & Mantis-Eval   \\ \midrule
Concatenate      & \textbf{59.0} & \textbf{51.6} & 22.4          & 25.1          & 55.7          & 38.7          \\
Pre-Cross-Attention  & 53.8          & 45.2          & \textbf{43.0} & 38.9          & 55.3          & 44.7          \\
Post-Cross-Attention & 48.9          & 40.9          & 38.3          & 37.0          & 54.0          & 47.0          \\
Hyper Attention      & 57.6          & 50.0          & 42.8          & \textbf{39.4} & \textbf{59.5} & \textbf{51.6} \\ \bottomrule
\end{tabular}
}
\caption{Comparison between different attention structure. Concatenate means direct concatenate visual and text feature sequences. We use \textbf{bold} to mark the highest score.}
\label{tab:cross}
\end{table}

\Cref{tab:cross} shows that the concatenate-based model which directly embeds image features into the input sequence of the language model, has the best performance in single-image understanding. On the other hand, utilizing Post-Cross-Attention results in the worst performance. Comparatively, Pre-Cross-Attention performs better but still incurs some performance loss. Hyper Attention, however, achieves comparable performance with concatenate-based model.

In evaluations involving videos and multiple images, we observe that the concatenate-based model may not follow textual instructions as accurately, leading to a significant performance degradation in multi-image scenarios. This is attributed to the inadequate training of inter-image attention, which significantly disrupts the model's hidden states. Conversely, both single images and multiple images share the same paradigm when performing cross attention with text, which allows its multi-image capability to be better generalized from single-image training. the Hyper Attention design stands out as particularly effective in balancing the model's capabilities for handling both single and multiple images, showcasing superior generalizability

We also explore the integration position of the hyper attention. As shown in \Cref{tab:indices}. The results indicate that even with just two layers of Hyper Attention, the model achieves impressive performance on single-image benchmarks, while also demonstrating generalization capabilities for videos and multiple images. However, when we apply a denser integration strategy by introducing eight layers of Hyper Attention, we find that it does not yield improved single-image performance at this scale of training data, and its zero-shot generalization is even worse. Therefore, we ultimately integrate only four layers into the entire model.

\begin{table}[tbp]
\centering
\resizebox{1\linewidth}{!}{
\begin{tabular}{@{}l|cccccc@{}}
\toprule
Hyper Attention Layers Indices     & GQA           & TextVQA       & MvBench       & VideoMME      & NLVR2         & Mantis-Eval   \\ \midrule
$[9, 27]$                & 55.1          & \textbf{51.3} & 42.2          & 38.2          & 58.3          & 48.4          \\
$[1,5,9,13,17,21,25,29]$ & 56.2          & 48.3          & 41.5          & \textbf{39.5} & 52.4          & 47.5          \\
$[1,9,17,25]$            & \textbf{57.6} & 50.0          & \textbf{42.8} & 39.4          & \textbf{59.5} & \textbf{51.6} \\ \bottomrule
\end{tabular}
}
\caption{Comparison between different layers for integrating hyper attention structures. We use \textbf{bold} to mark the highest score.}
\label{tab:indices}
\end{table}

\subsubsection{Design of Hyper Attention}

To further investigate the impact of the structural design of Hyper Attention on model performance, we start with a basic hyper attention model and gradually introduce adaptive gating, shared layernorm, and MI-Rope. The \Cref{tab:ablation} shows that, when incorporate adaptive gating, the single-image understanding performance is significantly improved. And if we use a shared layernorm, performance is further improved. In video scenario, we notice that even without any inter-image position encoding, the performance of video understanding is also improved, suggesting the temporality inherent in visual content can also be implicitly modeled by the model with the help of adaptive gating. However, when evaluating models with multiple images, the contextual position of the images is crucial and cannot be implicitly modeled. Therefore, it can be observed that incorporating adaptive gating and shared layernorm does not lead to performance improvement on multi-image benchmarks. However, with the introduction of MI-Rope, the metrics for various multi-image benchmarks have demonstrated significant improvement.

\begin{table}[]
\centering 
\resizebox{1\linewidth}{!}{
\begin{tabular}{@{}ccc|cccccc@{}}
\toprule
Adaptive Gating & Shared LayerNorm & MI-Rope    & GQA  & TextVQA & MvBench & VideoMME & NLVR2 & Mantis \\ \midrule
               &           &           & 53.3 & 44.6           & 40.2    & 38.1     & 52.7  & 41.9   \\
\checkmark      &            &            & 55.7 & 49.3           & 43.2    & 40.1     & 53.4  & 47.9   \\
\checkmark      & \checkmark &            & 58.1 & 49.7           & 42.8    & 38.4     & 54.9  & 46.1   \\
\checkmark      & \checkmark & \checkmark & 57.6 & 50.0           & 42.8    & 39.4     & 59.5  & 51.6   \\ \bottomrule
\end{tabular}
}
\caption{Ablation on the Adaptive Gating, Shared LayerNorm and MI-Rope.}
\label{tab:ablation}
\end{table}

\subsection{Distractor Resistance in Long Visual Contexts}
Recent works adopt the multimodal needle in a haystack~\citep{wang2024needle} approach to evaluate the understanding of long sequences. However, we notice that multimodal models, when understanding multiple images, are susceptible to interference from surrounding images, leading to visual illusions. The multimodal needle in a haystack evaluation cannot detect such errors. Therefore, we develop a challenge evaluation method to assess the distractor resistance of multimodal models in long visual contexts.


Specifically, we take samples from the MMBench dev set. For each test sample, we randomly select $N-1$ images from the original MMBench dev set as distractor and construct the model input in the format of \textit{Image 1: <|image|> Image 2: <|image|> ... Image N: <|image|>. In Image X, \{question\}}, where $N={1,5,10,20,50,100,200,400}$ and $X$ denotes the index of the image corresponding to the question. We use the CircularEval to measure the accuracy scores. For each question, we construct test samples with different orders of options and varying distractor images. The model needs to answer all test samples for a given question correctly for it to be counted as correct. Consequently, as the number of distractor images increases, the evaluation becomes significantly more challenging.

We compare \modelname with LLaVA-Next-Interleave 7B~\citep{li2024llava}, Mantis-Idefics2~\citep{jiang2024mantis}, Qwen-VL~\citep{Bai2023QwenVL} and mPLUG-Owl2~\citep{ye2024mplug}. LLaVA-Interleave-7B can handle approximately 20 images given 80GB of VRAM. By utilizing model parallelism, we extend its capacity for images to 50 images. However, LLaVA-Next-Interleave is unable to handle settings with more images. Mantis-Idefics2 can handle up to 100 images but costs 9 hours to finish the evaluation. 

\begin{figure*}[bt]
    \centering
    \includegraphics[width=\textwidth]{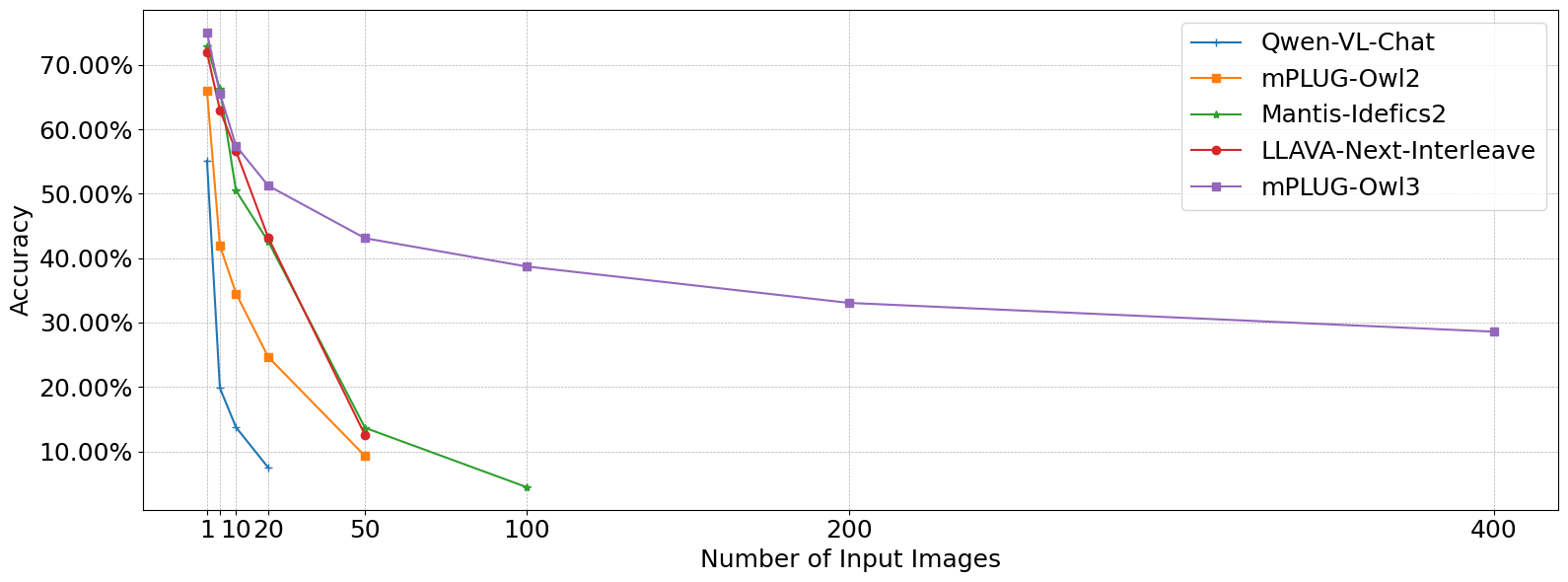}
    \caption{The performance of interference resistance with long visual context across LLaVA-Next-Interleave 7B~\citep{li2024llava}, Mantis-Idefics2~\citep{jiang2024mantis}, Qwen-VL~\citep{Bai2023QwenVL} mPLUG-Owl2~\citep{ye2024mplug}) and \modelname.}
    \label{fig:inter_res}
\end{figure*}

The results are shown in \Cref{fig:inter_res}. It can be observed that the introduction of distractor images results in a certain degree of performance loss for all the models. When the number of images reaches 20 and 50, the performance of LLaVA-Next-Interleave dramatically drops to 43.18\% and 12.52\%, respectively. We observe that when the number of images reaches 50, LLaVA struggles to consistently answer the questions accurately when different distractor images are present, resulting in a low accuracy rate. And when the number of images reaches 100, Mantis-Idefics2 fails to solve most of the problems correctly. In contrast, \modelname only drops to a performance level of 43.09\% when processing 50 images. As the number of images increases to 400, the performance of \modelname decreases to 28.58\%. Since our multi-image training data consists of only about 6-8 images, this also presents a challenge for our model. Nonetheless, \modelname can serve as a baseline for future research.

\subsection{Qualitative Results}
\modelname can handle various number of images and videos as inputs. In this section, we further investigate the ability of \modelname in real-world dialogue scenarios.

\subsubsection{Multi-Image Understanding}
\begin{figure*}[bt]
    \centering
    \includegraphics[width=\textwidth]{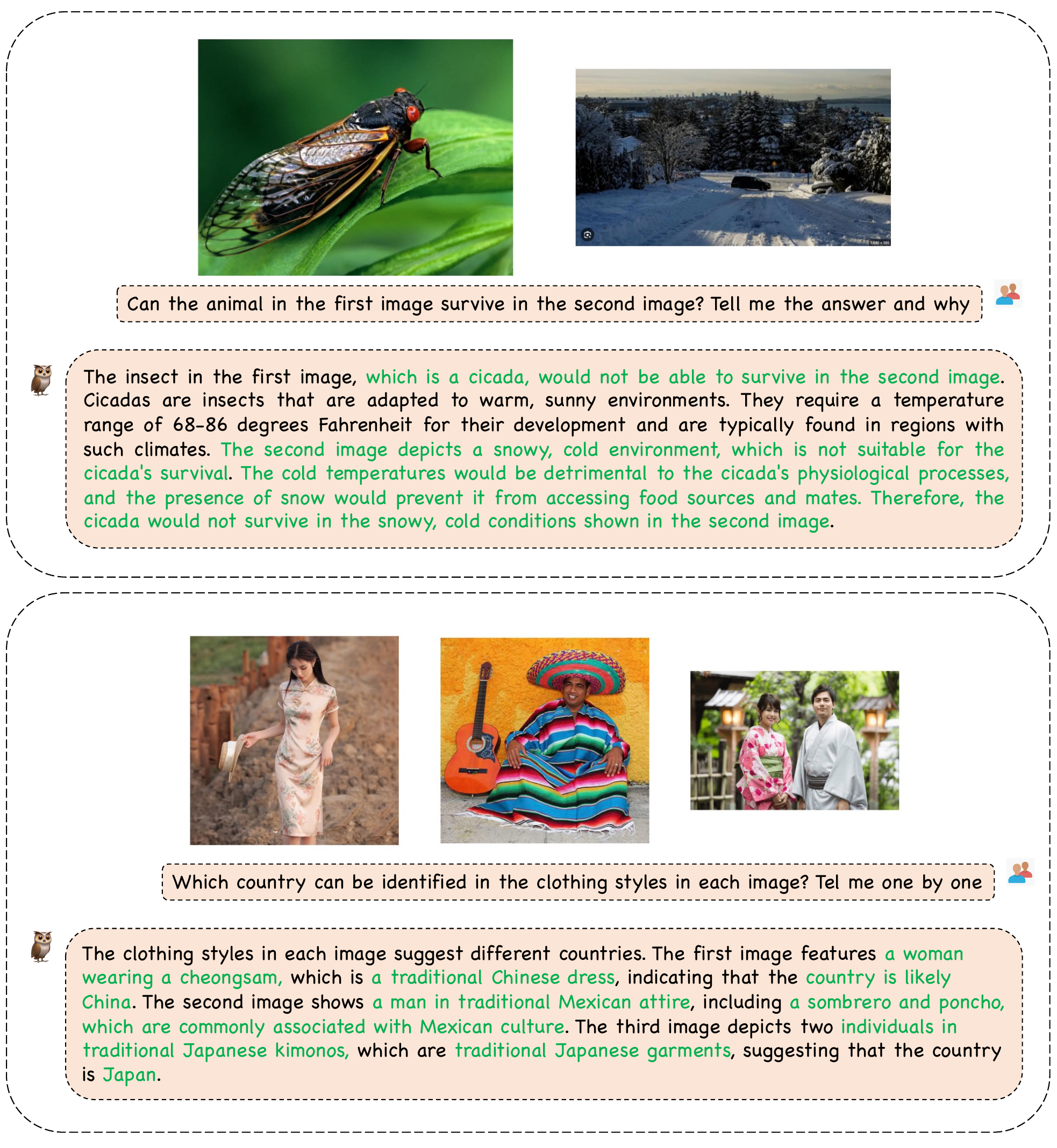}
    \caption{Examples for Multi-Image Understanding. We highlight the correct answers in green.}
    \label{fig:multi_1}
\end{figure*}

\modelname demonstrate state-of-the-art performance on multi-image understanding benchmarks. In this section, we present multi-image dialogue examples in real-world. In the first example shown in \Cref{fig:multi_1}, it can be observed that \modelname can activate the knowledge it learned based on the content of the images and perform cross-image reasoning. The second example demonstrates that the model can accurately distinguish the content of multiple images and respond appropriately based on cultural knowledge.

\begin{figure*}[bt]
    \centering
    \includegraphics[width=\textwidth]{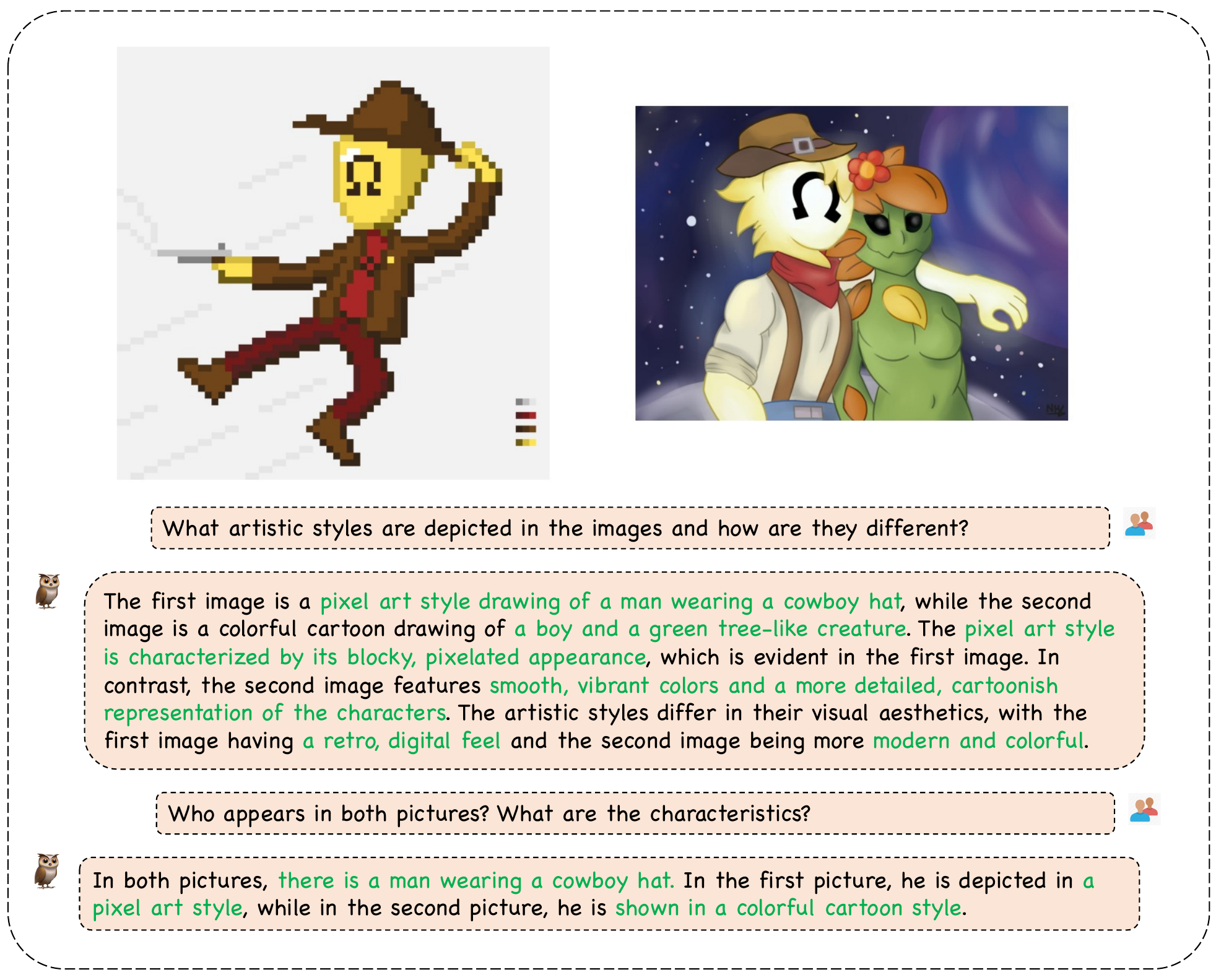}
    \caption{Examples for Multi-turn Multi-Image Dialogue. We highlight the correct answers in green.}
    \label{fig:multi_2}
\end{figure*}

\Cref{fig:multi_2} shows a multi-turn dialogue example. \modelname can find the differences between two images in various views. Besides, it can describe the correlations between images.

\subsubsection{Video Understanding}
\begin{figure*}[bt]
    \centering
    \includegraphics[width=\textwidth]{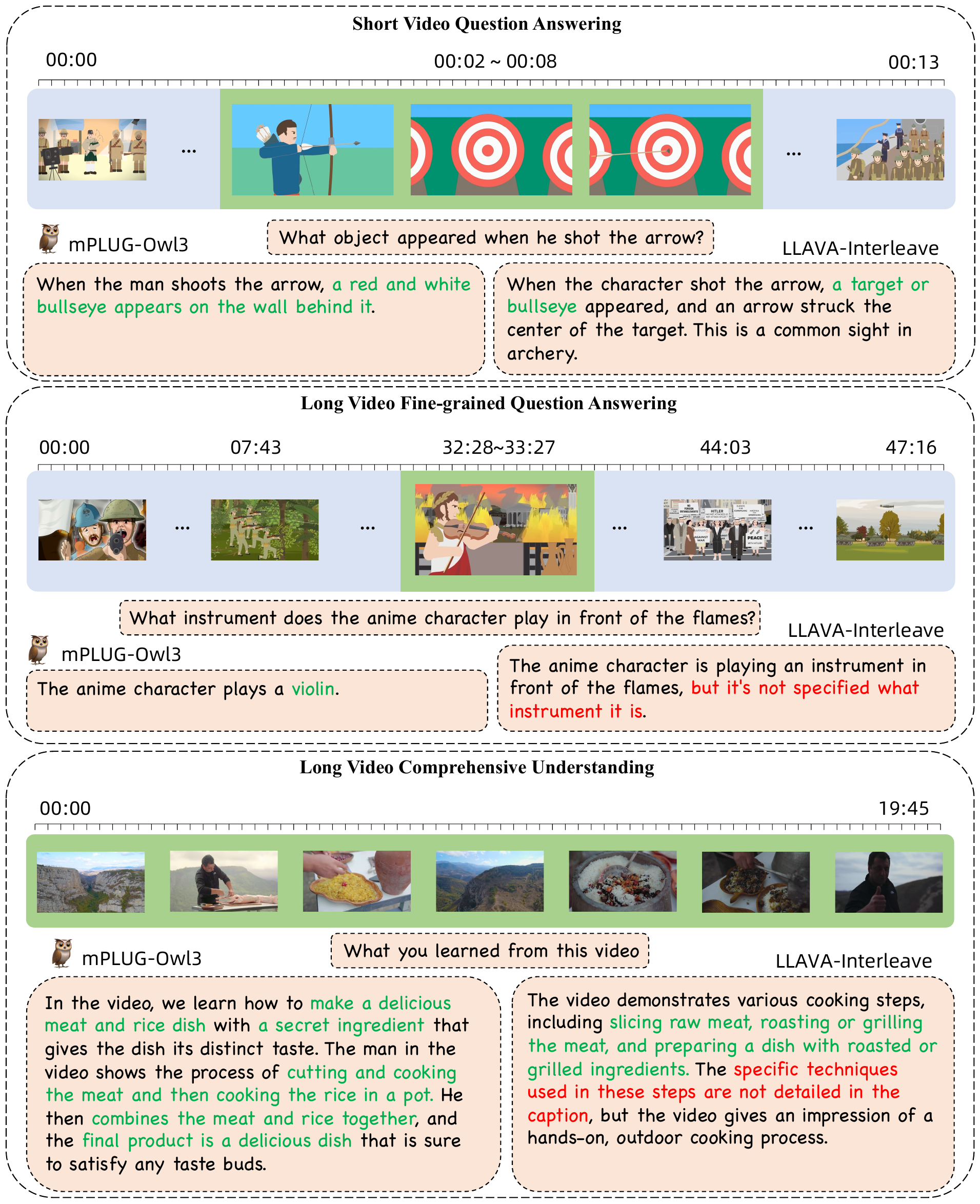}
    \caption{Comparison between \modelname and LLaVA-Interleave across Short Video Question Answering, Long Video Fine-grained Question Answering, and Long Video Comprehensive Understanding. We highlight the correct and relevant parts of the answers in green, while the parts that fail to answer the question correctly are marked in red. Additionally, the segments of the video that are relevant to the questions are highlighted with a green background.}
    \label{fig:video_case_cat}
\end{figure*}

We showcase the video understanding capabilities of \modelname. First, we compare it with LLaVA-Next-Interleave in Short Video Question Answering, Long Video Fine-grained Question Answering, and Long Video Comprehensive Understanding. For LLaVA-Next-Interleave, we input 8 frames, while for \modelname, we input 128 frames, which are the maximum numbers of images that can be accommodated by the two models on a V100-32G. The samples are shown in \Cref{fig:video_case_cat}.

In the short video tests, both LLaVA and \modelname can provide correct answers. \modelname tends to describe the attributes of objects based on the actual content seen. In long video lasting more than 40 minutes, when we ask about a specific detail, LLaVA fails to handle the long sequence and loses fine-grained information, rendering it unable to provide accurate information. On the other hand, \modelname accurately captures key segment information within a long video. Additional, we have both models summarize the content of a longer video. \modelname's response is very detailed, not only providing an overall summary but also introducing the process in order. LLaVA-Next-Interleave’s response, however, is more general and lacks detail. The comparative results indicate that \modelname not only efficiently encodes long visual sequences but also captures and effectively utilizes both global and local information.

\begin{figure*}[bt]
    \centering
    \includegraphics[width=\textwidth]{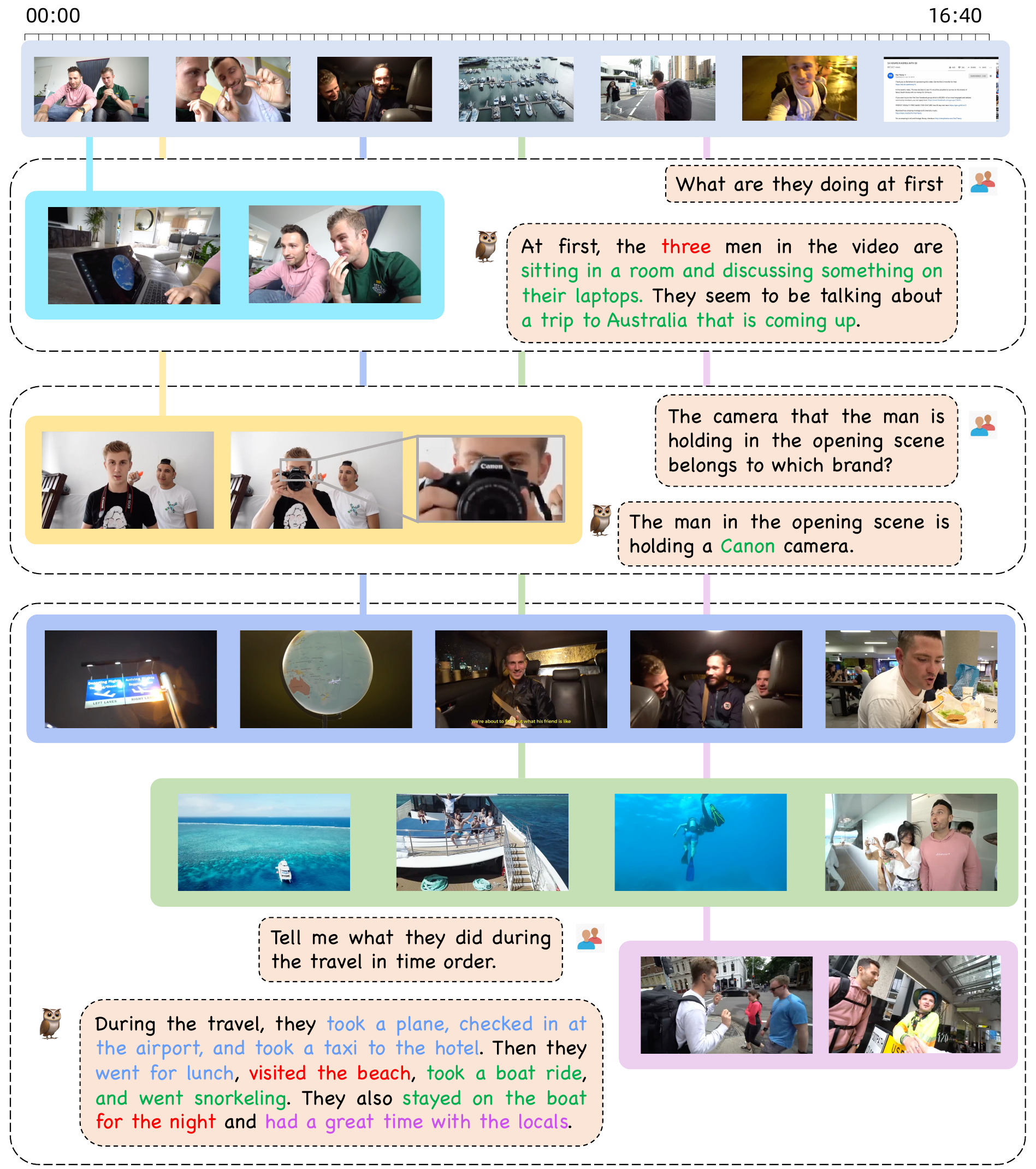}
    \caption{Examples of mPLUG-Owl3's understanding of complex video content}
    \label{fig:video_case_one}
\end{figure*}

We also test \modelname in multiple rounds using a long video that featuring many scenes. For clarity, we place the relevant segments beside the dialogue in the figure. During the test, we input only the complete video to the model. The dialogue is shown in \Cref{fig:video_case_one}.

First, we ask a question with a temporal constraint, and \modelname accurately understands the concept of "at first" and correctly describes the detail of ``sitting in a room and discussing something on their laptops.'' However, the response incorrectly counts the number of people. The segment has only two people. We find that the model is confused by a later scene involving more people. We also notice that the visual content of this segment does not involve Australia as a destination, but the model can infer this from some diagrams later in the video, which makes the response more detailed. Then, we ask about the camera brand in a frame that briefly appears, and \modelname accurately notices the "Canon" logo in the image and provides the correct answer. Finally, we ask the model to describe the travel in order of time. We use the same color to identify the content described by the model and the corresponding video segments. Since the video involves many scenes and events, this poses a great challenge to the model. It can be observed that \modelname accurately details the travel according to the timeline of the video. However, we also notice that \modelname exhibits some hallucinations, incorrectly interpreting the reefs captured in the video as a beach. Additionally, while the activities on the boat happen during the day, \modelname, influenced by other nighttime scenes, makes an incorrect statement.

\section{Related Work}
\subsection{Multimodal Large Language Models}
With the development of large language models (LLMs), researchers are exploring the integration of vision and other modalities into LLMs.
These multimodal large language models (MLLMs) can perceive visual contents, conduct visual reasoning, and engage in multimodal dialogue with humans.

Based on the way visual features are integrated into language models, MLLMs can be divided into three categories: 
\begin{itemize}
    \item Models like LLaVA~\citep{Liu2023Llava} and CogVLM~\citep{Wang2023CogVLMVE} use an MLP to map visual features into the representation space of the language model, and directly concatenate them with the text sequence.
    DeepSeek-VL~\citep{lu2024deepseek} employs multiple visual encoders to obtain richer visual representations. 
    While these methods can preserve fine-grained visual information, they consume a large number of tokens which slows down both training and inference.
    \item To reduce the number of tokens, Mini-GPT4~\citep{Zhu2023MiniGPT4}, mPLUG-Owl~\citep{ye2023mplugowl}, and Qwen-VL~\citep{Bai2023QwenVL} adopt a structure similar to Q-Former~\citep{Li2023BLIP2}, compressing the token count to a fixed size through learnable queries and cross-attention with visual features. InternLM-XComposer~\citep{zhang2023internlm} and IDEFICS2~\citep{laurenccon2024matters} also use the similar method. Models like InternVL~\citep{chen2024internvl} and InternLM-XComposer-2.5~\citep{zhang2024internlm} use patch merge to compress visual tokens by several times. MiniGemini~\citep{li2024mini} uses a low-resolution visual representation as a query to compress and aggregate high-resolution visual features through cross-attention. These methods can reduce the number of tokens but all suffer from information loss.
    \item Flamingo~\citep{alayrac2022flamingo} first proposed embedding cross-attention layers into the language model, integrating visual features into the intermediate representations of the language model. IDEFICS~\citep{laurencon2023idefics} and EVLM~\citep{chen2024evlm} have also trained MLLMs based on this structure. This method avoids occupying the context window of the LLM, saving computational overhead. However, it introduces more parameters and may interfere with the intermediate representations of the pre-trained language models, making the performance of such models often sub-optimal compared to mainstream models.

\end{itemize}
mPLUG-Owl3 maintains the raw visual features during the multimodal fusion to prevent the information losing. Besides, we propose a light weight module named Hyper Attention to perform cross-attention and self-attention in parallel inside the language models. By sparsely replacing several of the transformer blocks in the Large Language Model with Hyper attention blocks, mPLUG-Owl3 can balance model performance and inference efficiency, achieving state-of-the-art performance in single-image, multi-image, and video understanding, and its inference efficiency far exceeds that of existing models.
\subsection{Multimodal Models with Interleaved Support}
Early-stage models, trained exclusively on single-image inputs, exhibit limitations in image-text interleaved scenario. Recent research are expanding the capabilities of multimodal models to process multiple images inputs.

\begin{itemize}
\item Video is a special form of multi-image existence, and MLLMs related to video understanding treat frames as multiple images with temporal correlation as input. VideoChat2~\citep{li2023videochat} propose a Global Multi-Head Relation Aggregator to perform temporal message passing and use a Q-former to adapt the feature of video frames into language model. VideoLLaMA2~\citep{cheng2024videollama} not only reads images but also expands the model's audio comprehension capabilities, ensuring that the information in the video is fully utilized. ShareGPT4Video~\citep{chen2024sharegpt4video} propose to improve the video understanding by introducing GPT-4 annotated video caption as pretrain data. 

\item In general multimodal dialogue, the model needs to have a more general multi-image understanding capability, including in-context learning, cross image reference, comparison, and reasoning. Flamingo~\citep{alayrac2022flamingo} demonstrates limited in-context learning capabilities, while Idefics2~\citep{laurenccon2024matters} has acquired a broader multi-image understanding ability through multi-image training data. Mantis~\citep{jiang2024mantis} and LLAVA-Interleave~\citep{li2024llava} further enhance the model's multi-image understanding capabilities by constructing more refined multi-image understanding datasets.
\end{itemize}

\modelname abandons the approach of concatenate visual features to text sequences and instead employs efficient Hyper Attention for multimodal interaction. This not only enhances its capability for understanding multiple images and videos, but also enables it to handle very long visual sequence inputs with low resource overhead.
\section{Conclusion}
In this paper, we present \modelname, a multi-modal large language model that significantly advances the state-of-the-art in handling both single-image, multi-image and video tasks. The introduction of novel Hyper Attention enables the \modelname to maintain the fine-grained visual input and effectively fuse visual and textual information, leading to superior performance across various benchmarks. We also propose a challenging long visual sequence evaluation named Distractor Resistance. Notably, \modelname excels in managing ultra-long visual sequences and demonstrates a strong performance in evaluation. We believe that \modelname reveals a direction for building an efficient and effective multi-modal large language model. We hope it can become the foundation for future research.

\bibliography{iclr2024_conference}
\bibliographystyle{iclr2024_conference}

\end{document}